\documentclass[aps,pra,twocolumn,english]{revtex4}
\usepackage[T1]{fontenc}
\usepackage[latin9]{inputenc}
\usepackage{babel}
\usepackage{graphicx,amsmath,amssymb,color}
\usepackage[normalem]{ulem}
\usepackage{amsfonts}
\usepackage[toc,page]{appendix}
\usepackage{hyperref}
\usepackage{latexsym}
\usepackage{amsfonts}
\usepackage{algpseudocode}
\usepackage{amsthm}
\usepackage{mathrsfs}
\usepackage{natbib}
\usepackage{color,verbatim}
\usepackage{psfrag}
\usepackage[svgnames]{xcolor}
\usepackage{tikz}

\usepackage{multirow}

\usepackage[textsize=footnotesize,backgroundcolor=red!25]{todonotes} 

\begin{document}

\title{Toward a standardized methodology for constructing quantum computing use cases}
\author{Nicholas Chancellor$^\ddagger$, Robert Cumming$^\dagger$, Tim Thomas$^\dagger$}
\email{nicholas.chancellor@gmail.com}
\address{$\dagger$ AppliedQubit, London, UK \\
$\ddagger$ Department of Physics and Durham Newcastle Joint Quantum Centre\\ Durham University, South Road, Durham, UK \\
}

\begin{abstract}
We propose a standardized methodology for developing and evaluating use cases for quantum computers and quantum inspired methods. This methodology consists of a standardized set of questions which should be asked to determine how and indeed if, near term quantum computing can play a role in a given application. Developing such a set of questions is important because it allows different use cases to be evaluated in a fair and objective way, rather than considering each case on an ad hoc basis which could lead to an evaluation which focuses on positives of a use case, while ignoring weaknesses. To demonstrate our methodology we apply it to a concrete use case, ambulance dispatch, and find that there are some ways in which near term quantum computing could be deployed sensibly, but also demonstrate some cases ways in which its use would not be advised. The purpose of this paper is to initiate a dialogue within the community of quantum computing scientists and potential end users on what questions should be asked when developing real world use cases.
\end{abstract}

\maketitle

\section{Introduction}

Quantum computing is rapidly advancing as a field and has arrived at the point where we need to ask questions about what the earliest use cases should look like \cite{Biswas2016a,Mohseni2017a}. While there have been many proof-of-concept studies on how quantum computers could be used in a diverse set of fields including scheduling \cite{Venturelli15a}, vehicle routing \cite{crispin13a,Feld19a}, air traffic control \cite{Stollenwerk19a}, optimising radar waveforms \cite{coxson14a}, seismology \cite{Souza20a},  hydrology \cite{omalley18a}, and finance \cite{Orus18a,marzec16a} relatively little attention has been paid to the question of how to identify whether an application is promising in the near term. 

There has been recent work on the related issue of determining whether specific problem instances are suitable for quantum computing, in particular relating to application of an algorithm known as the quantum approximate optimisation algorithm (QAOA) to a class of max-cut problems \cite{Moussa20a}, however this work takes a much narrower scope than our current work. In the very long term, almost any application which involves solving computationally difficult problems is a potential target for quantum computing, making a case for quantum computing is simply a matter of developing an algorithm, or adapting one which already exists. In the short to medium term however, this question is much more nuanced, as machines can be expected which are on one hand potentially very powerful, but also very limited. The quantum processors available during this time period are sometimes referred to as noisy intermediate scale, or NISQ processors \cite{Preskill18a}.

Algorithms in this are likely to be hybrid quantum/classical algorithms, rather than running entriely on a quantum machine. For gate model quantum computing, such algorithms such as QAOA \cite{Farhi14a,Hadfield17a} which we have mentioned previously and variatonal algorithms such as the variational quantum eigensolver (VQE) \cite{McClean16a,Kandala17a}. For continuous time approaches, such as quantum annealing, there are a variety of hybrid tools which allow classical guesses to be incorporated into algorithms, including the reverse annealing feature implemented on the devices produced by D-Wave Systems Inc.~\cite{chancellor17b,reverse_anneling_whitepaper} as well as a variety of proposed methods \cite{Perdomo-Ortiz11,Duan2013a,Grass19a}. 

Finding highly promising applications for these machines is far from trivial, and is also, for the most part, not actually a `quantum' problem, but rather one of finding applications which are the right shape and size for quantum computers. An additional challenge is maintaining objectivity when evaluating these use cases, it is all too easy to focus on the positives of a quantum use case, while ignoring aspects which are potentially problematic. The best way to remain objective in these evaluations is to develop standardized criteria, which are applied to every potential use case in a systematic way, rather than taking a fresh ad hoc approach to every new case.

Evaluating the potential of a use case for quantum computing is necessarily a nuanced process, which depends on how the quantum computer is deployed. This is crucial, especially when considering factors around whether or not the relatively small size of early quantum computers will be a major hindrance. For this reason, the task of deciding whether a use case is promising for quantum computing is intimately tied to the \textbf{deployment} of the quantum computer, in other words, which subproblems it will be given to solve and how it will be used. We therefore propose a methodology for iteratively determining a suitable deployment of quantum computing for a given problem.  We also include the possibility that a use case is not suitable for quantum computing but is for quantum inspired algorithms \cite{Montiel20a,Snelling20a,Crosson16a}, which are not subject to the same restrictions.

\begin{figure*}
 \begin{centering}
 \includegraphics[width= 10cm]{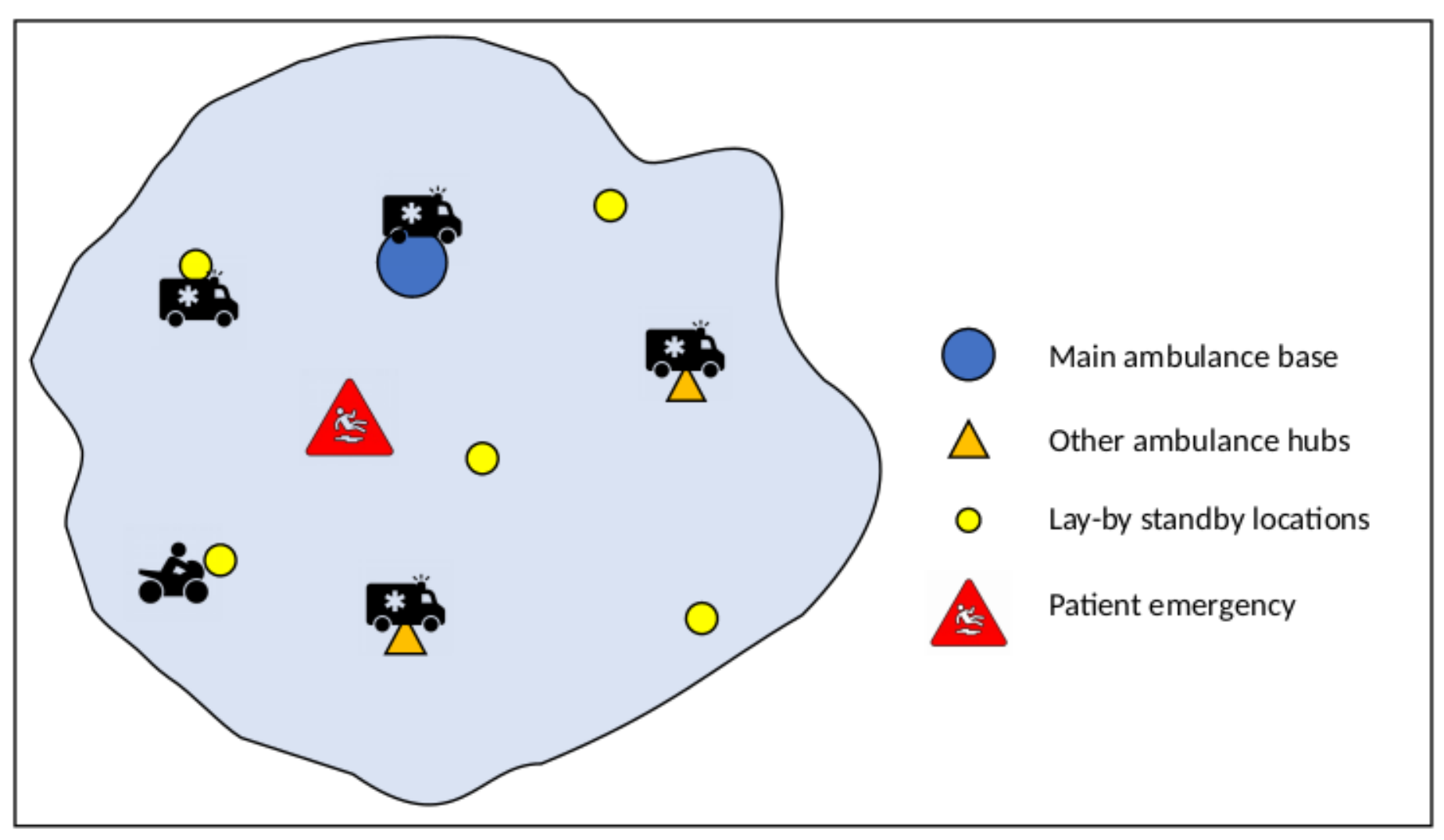}
 \end{centering}
 \caption{Cartoon illustration of the ambulance dispatch problem, with hubs and other locations where ambulances and related resources can be stationed. \label{fig:ambulance_illustration}}
 \end{figure*}

To keep our methodology grounded and to demonstrate its usefulness, we apply it to a real world use case, the ambulance dispatch problem. The ambulance dispatch problem as we define it is the problem of deciding how to deploy a variety of assets ranging from full scale ambulances to motorcycles to emergencies which may arise, and to assign them locations where they are prepared to respond to emergencies as necessary so that an area is effectively covered. An example of the type of situation under consideration is illustrated in Figure \ref{fig:ambulance_illustration}.  It depicts a territory covered by an ambulance service which has a main base location, 2 other operational hubs and a number of other locations where ambulance resources can be located.  A new incident requiring attendance from the ambulance service is shown in the red triangle.  There are 4 standard ambulances available to attend based at various locations as shown and in addition a paramedic on a motorcycle is also available.  The challenge is which resource should be dispatched.  As a particular instance this might be considered to be straightforward to address, once the problem is extended to include multiple incidents and an available resource of ${\sim}100$ units, it quickly becomes intractable to solve optimally using classical computing capabilties.

The ambulance dispatch problem is ideal to demonstrate our methodology because it is complex, multiple approaches are conceivable, and there are a variety of potential subproblems which a quantum computer could be called upon to solve. Phrased another way, it is not a problem which is already written as a computer science problem, but is a raw, real world problem which has not yet been simplified and reduced to mathematics. 

As we see later, there are at least two potential approaches where quantum computing could help with this problem, the first is to directly optimise over an objective function related to the quality of service, and ability to respond, the second is to construct probabilistic graphical models \cite{KollerBook} and optimise over probability, effectively simulating conditioned on rare events.

For the purposes of clear communication, we define some terms which we will use in this manuscript:

\begin{itemize}
    \item \textbf{Use case}
    The application for which quantum (inspired) algorithms are being examined, this is the context of the overall goal. For example ambulance dispatch is a use case were the goal is to improve the performance of the ambulance service by allocating resources more efficiently. A use case is broader than a single specific objective function.
    \item \textbf{Deployment}
    A specific proposal for how to apply quantum or quantum inspired algorithms to a given use case. The deployment includes how the computational tools will be used to get improved results against the use case. This paper focuses on both deciding the quality of individual deployments, and developing improved deployments.
    \item \textbf{Action}
    A decision on what to do with the deployment, for example proceeding in developing a specific deployment using quantum computing is an action. Most actions can be considered as final states of the decision methodology except for reformulation of the depolyment.
    \item \textbf{Reformulate}
    A special action where the deployment is redesigned to address a decision point under which it was previously found unsuitable.
    \item \textbf{Decision point}
    A question which is asked about the current deployment to decide how to proceed, either to reach an action to take for the deployment or to proceed to another decision point.
\end{itemize}

In section \ref{sec:useDecide} we introduce a simplified methodology for solving a problem which we call the \textbf{deployment decision problem}, which is to decide the validity of a use case with a fixed deployment of quantum resources, and to determine if quantum inspired may be suitable instead. In the next section, section \ref{sec:iterDeploy}, we no longer require a fixed deployment, and propose a formalism to iteratively change the deployment to improve a use case, we refer to this problem as the \textbf{deployment design and decision problem} problem. In section \ref{sec:ambulanceExamp} we demonstrate our iterative methodology being applied to the ambulance dispatch problem. Finally, we conclude with some discussion and final remarks.

\section{The use case decision problem \label{sec:useDecide}}

The first problem we consider is what we refer to as the \textbf{deployment decision problem}, in other words, given a potential use case, including how the quantum computer will be deployed, how do we decide if it is suitable for quantum computing. To make this decision, we consider a number of important factors. An important distinction here is the difference between the overall problem, and individual instances where the solution is deployed; for example, if quantum methods are used to train classical artificial neural networks, then the solution can potentially be reused an unlimited number of times. Throughout this section we refer to the problem overall and also to individual instances of the problem; in this case each `instance' refers to an individual problem the quantum processor solves.

\subsection{Economic or social value}

How much value would a better solution to the problem give? This could either be economic value, increasing revenues or reducing costs (or some combination), or social value i.e.~more efficient ways to deal with emergencies, better functioning of public infrastructure, etc.

\subsubsection{Overall value \label{point:overall_value}}

If an unlimited number of better solutions to this problem were available, how much value would be generated?

\subsubsection{Per instance value  \label{point:instance_value}}

How often does this problem need to be solved, how much value can be derived from solving the most valuable single instances? Can a solution for a single instance be reused?

\subsection{Limitations, requirements, and constraints}

What limitations exist? What are the constraints? Note that these apply to quantum (inspired) and classical equally, so some constraints may make an application more appealing for quantum (inspired) methods. These limitations could include for instance a subset of conditions which  must be met for a solution to be interesting, or a minimum size of problem which must be solved to be relevant to the end user.
\subsubsection{Time Constraints \label{point:time}}

Is a solution needed within a certain time frame? If so, what what is this time frame?

\subsubsection{Information/communication constraints \label{point:comms_info}}

Does the problem require information which cannot be practically communicated to a central location, or does it need to be solved `locally'? Cases where a problem does not have to be solved locally include machine learning approaches to problems where a quantum machine is used to train a neural network, but the network itself is not quantum.

\subsubsection{Data throughput/ instance size \label{point:throughput}}

Does the application require a large throughput of data? What instance sizes are needed to be provided to the quantum or quantum inspired part of the algorithm when solving the problem? 

\subsection{Current methods}

Information about the algorithms/techniques which are currently used, in particular how well they perform and whether there is actually room for quantum or quantum inspired methods to bring meaningful improvement.

\subsubsection{Computational hardness \label{point:hardness}}

Does the problem appear to belong to a class of problems which are difficult classically, for instance does it reduce to an NP-hard problem?

\subsubsection{Optimality of current methods \label{point:current_methods}}

Is there an obvious way to improve the current methods without invoking quantum (inspired) techniques?

\subsubsection{Solver type \label{point:solver_type}}

 Are the solvers which are used of a type already known to map to quantum hardware?
 
 \subsection*{Decision protocol}
 
\begin{figure*}
 \begin{centering}
 \includegraphics[width= 10cm]{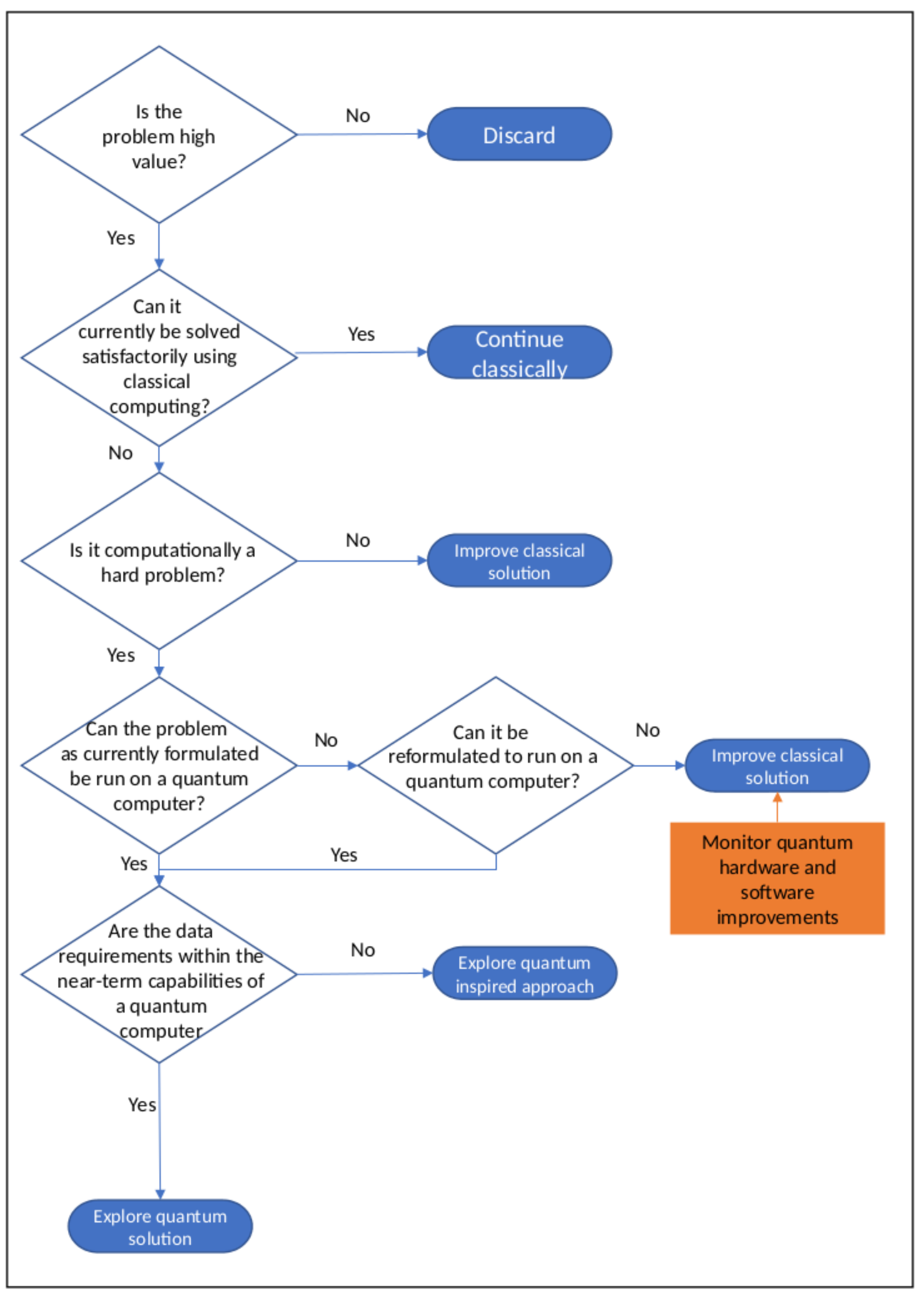}
 \end{centering}
 \caption{Decision protocol for a quantum computing use case. \label{fig:decision_diagram}}
 \end{figure*}

 Based on these considerations, a decision protocol is given in fig.~\ref{fig:decision_diagram} which gives questions which should be asked when considering a use case for quantum computing. While this decision process has the advantage of being relatively straightforward, it is (formally speaking) a directed acyclic graph (paths may reconnect, but no way to travel in a loop), which therefore means that many of the aspects of the usage of the quantum computer (which we refer to as the deployment) are treated as fixed. Since the same decision points used in this section will be used in the iterative methodology in section \ref{sec:iterDeploy}, we reserve the detailed discussion of the points for that section.

 This method is ideal if the end user already has a well formulated idea of how the quantum computer will be deployed and just has questions about whether it should, or how it can be encoded into a quantum computing device. However, in many cases whether or not a use case is viable depends on the deployment. As a very simplified example, consider an (obviously flawed) deployment of the ambulance dispatch problem where a quantum computer is placed aboard each ambulance. This simplified example fails the decision method given in fig.~\ref{fig:decision_diagram} at the first decision point, the problem of deciding where an individual ambulance should go is not high enough value to justify the use of a (high cost) quantum computer. On the other hand, a different deployment, where the quantum computer is used centrally e.g. accessed on a cloud basis, would easily pass the first decision point, as better routing of ambulances throughout a city is valuable enough to justify the use of a quantum computer. 
 
 Another example relates to what subproblem the quantum computer is deployed to solve, finding the optimal route for a single ambulance between two points is a computationally easy problem. Even if the quantum computer were deployed centrally, this deployment of quantum computing to the ambulance dispatch problem would fail on the second or third decision point (depending on whether this is a bottleneck classically). On the other hand, there are many hard subproblems related to routing between multiple points, or statistical inference of future behaviour, which would pass both of these decision points.
 
 Given that often the question we really want to answer is not just whether a particular deployment of quantum computers makes a good use case, but rather what is the optimal deployment, we consider a more sophisticated iterative protocol in the next section, which allows not only for reformulation of the problem given to the quantum computer itself, but also reformulation of the deployment.

\section{Iterative design methodology including deployment\label{sec:iterDeploy}}

We now consider a design process where the deployment can be changed iteratively, which we call the \textbf{deployment design and decision problem}; this is fundamentally a different problem than the use case decision problem. This process also starts with a fixed deployment, but includes the possibility that the deployment can be updated if the protocol fails. If there is ever a case where it is no longer possible to alter the deployment to cure potential issues, then the use case is abandoned. This protocol also has the possibility of distinguishing between use cases which are suitable for quantum inspired algorithms and those which are not. The possible actions which are allowed in this protocol are as follow:

\subsection*{Decide use case is not suitable}

The use case is not well suited for either quantum or quantum inspired algorithms, and it is unlikely that related use cases can be found. This will usually come about if the overall value is too low, the problem is already trivial to solve with classical methods, other basic conditions are not met such as attempting to solve  a problem which does not have sufficient social or monetary value, or all options for deploying quantum (inspired) methods have been exhausted.

\subsection*{Explore quantum and quantum inspired algorithms}

The use case and deployment are suitable to try to design algorithms which directly use quantum subroutine calls, examine the current algorithms carefully and see where a quantum subroutine might help. Use cases for genuinely quantum devices will almost always also be suitable for quantum inspired.

\subsection*{Explore quantum inspired algorithms only}

This deployment is appealing, but some unavoidable constraints make the application not suitable for near term quantum hardware. This could for example come from an unavoidable large data throughput, or a problem which must be solved locally in a location where installing a quantum computer would not be practical or economical. Note that while these are not appealing use cases for \emph{near term} quantum computers, it is likely that they would become appealing for a fully quantum approach with more advanced hardware.

\subsection*{Reformulate problem}

The current deployment is not suitable for quantum computing, but a different deployment may be. This could be for example a case where a `meta' problem is considered, for instance use quantum (inspired) algorithms to formulate rules on which a local classical system makes decisions. This could include machine learning techniques, but is not limited to them. After reformulation, all of  the context around the use case except for overall value will change and should be re-examined. If all reasonable formulations have been explored and none of them have been found suitable for quantum (inspired) algorithms, then discard.

\begin{figure*}
 \begin{centering}
 \includegraphics[width= 10cm]{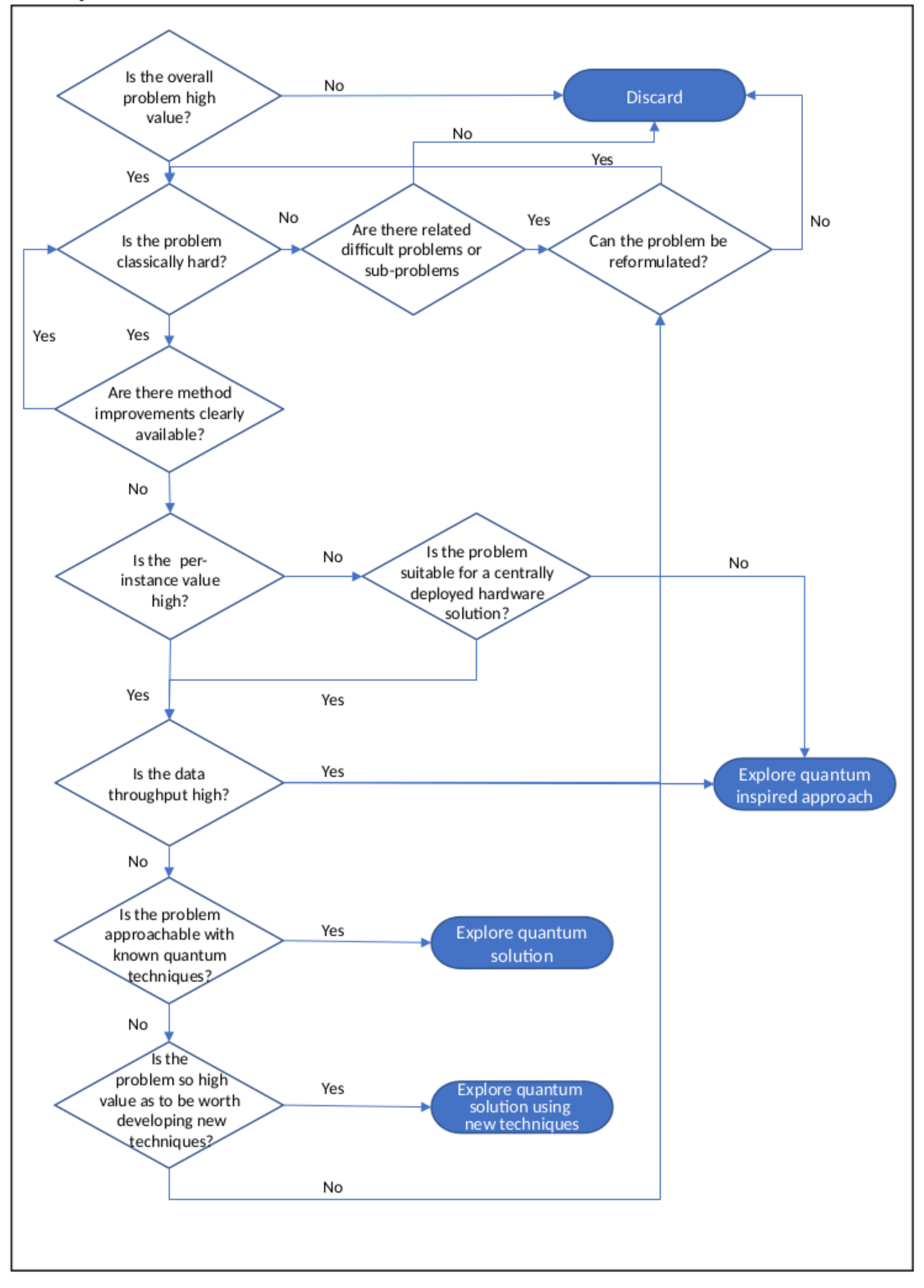}
 \end{centering}
\caption{Deployment design and decision protocol represented as a flowchart. \label{fig:design_flowchart}}
 \end{figure*}
 
 \subsection*{Decision points}
 
 Given the iterative nature of the protocol considered here, the decision protocol necessarily has cycles, as depicted in Fig.~\ref{fig:design_flowchart}.

 Also, unlike the simpler decision making flowchart, this flowchart includes multiple outgoing edges from some boxes, in these cases both routes are valid, and the choice of which to pursue (or to pursue both in parallel) should depend on the priorities of the person applying the protocol. The most common bifurcation of this type is when deciding whether to develop a quantum inspired solution or to try to reformulate the use case to make it suitable for genuinely quantum computers.
 
While many of the decision points were partially covered previously in the simpler decision making framework, we set out the full set of decision points for this iterative design methodology as follows:

 \begin{itemize}
 \item[1]Overall value: Does a better solution to the problem promise to deliver high economic or social value? This is the only element of the decision process which cannot be re-visited since it cannot be changed by reformulation.
 \item[2]Classical hardness: Is the subproblem being solved by quantum (inspired) hardware difficult with current classical methods? If not, then it is likely not worth the effort to implement quantum or quantum inspired algorithms. In either case, there are further considerations:
 \begin{itemize}
 \item[2a]Current methods: Are there obvious improvements to the current classical methods? Reflect on whether the currently used algorithms can be obviously improved - this is an important filter for problems where classical solution methods have not been very extensively explored.
 \item[2b]Other difficult subproblems: Has the use deployment been poorly formulated or are there simply not difficult subproblems? This decision point is used to catch use case proposals which simply have poor deployment, where the subproblem assigned to a quantum (inspired) processor is not appropriate. An example is a problem which is too large to feasibly be solved on a quantum computer, for example directly optimising all ambulance routing for a large city.

 \end{itemize}
 \item[3]Instance value: Does each instance of the subproblem give a high enough value to justify a quantum approach? This is to prevent use cases where many expensive quantum computers need to be deployed in ways which are not practical.
 \begin{itemize}
 \item[3a]Central server?: Can the instances be solved centrally? If communication with a central server is possible, then even if individual instances are low value they can still be solved centrally. This includes situations where quantum computers are used to train machine learning algorithms, which can be deployed classically.
 \end{itemize}
 \item[4]Data throughput: How large are the instances which need to be solved on the quantum processor, and how many are there? While quantum computers are very promising, near term ones will be best suited for small difficult subproblems. Consideration also needs to be given to how many of these problems must be solved to deliver the solution, and whether this is feasible.
\item[5]Approachable with known quantum techniques: Can currently known techniques be used to approach the problem? This does not mean that the entire mapping needs to be known in detail, but rather asks if there are a set of known techniques which can plausibly solve the class of problem which is being proposed for the quantum computer.
\item[6]Overall value: Does a better solution to the problem promise to deliver \emph{very} high economic or social value? If no technique is known to solve the problem using quantum computing, it may be worth developing new bespoke techniques, but only if the value is very high.
 \end{itemize}

\section{Example: ambulance dispatch problem\label{sec:ambulanceExamp}}

To demonstrate the advantage of having a formal decision making methodology for specific deployments of quantum computing related to a given use case, we consider the ambulance dispatch problem. The ambulance dispatch problem is the problem of how to dispatch different kinds of assets which a health service has at its disposal, which may include full ambulances, motorbikes, cars, and other vehicles. Multiple objectives can be considered, for example the minimization of wait times, optimal coverage of areas, ability to respond to crisis events, etc. The ultimate solution to this problem is decisions on where assets should locate themselves and which should respond to calls. While making all of these decisions on a quantum computer is not feasible, any tools which can help with these decisions should also be considered as partial solutions to the problem.

 As with most complex use cases, there are many different subproblems which could potentially be solved by quantum or quantum inspired methods, and therefore many potential deployments of these techniques. Because of the complexities of this use case, it is an ideal setting to demonstrate the advantages of having an established design methodology with fixed decision points. Here we consider several different potential deployments of quantum or quantum inspired algorithms.
 
 Before discussing the individual deployments, we consider decision point one in Fig.~\ref{fig:design_flowchart} since it does not depend on the details of the deployments. We find that the ambulance dispatch problem clearly has a high social value since it can save lives, furthermore, it may simultaneously have significant economic value if it reduces the total number of ambulances (and crews) to deliver the required level of service performance.

\subsection*{Deployment 1: Full routing plans by direct optimisation}

As the first example, let us consider a simple and direct approach, which is to assign a utility function to each possible location of an asset based on expected utility of having the asset at a given location, and subject to constraints.

If we consider the interactions between ambulance assets (point 2 in Fig.~\ref{fig:design_flowchart}), in particular the fact that the utility of having an asset in an area will be less if the area already has assets of the same type nearby then the solution space size will grow exponentially with the number of assets and possible locations they could be placed. This problem is likely to be hard to solve classically, at least at busy times when the problem becomes complex, and there are not obvious improvements to classical optimisation methods which would render these problems trivial.

Considering the next point, point 3, the value of solving each instance will be relatively low, and deploying a quantum computer in each ambulance would probably be impractical. However, most of the assets will be in almost constant communication with the central dispatcher, so a central server co-located with the dispatcher or used as a cloud service by the dispatcher could easily be possible. 

Moving on to point 4, the full routing plan for ambulance resources within a large region will probably be data intensive, and so trying to pass the entire problem of planning the routing to a near term quantum computer is unlikely to be feasible simply because the computer itself will not be large enough. There is however scope for quantum inspired algorithms, which do not share the same limitations, or for fully quantum computers in the longer term. The path through the flowchart which leads to this decision is summarized in Table \ref{tab:full_direct_opt}.

\begin{table}
\begin{tabular}{|c|c|}
\hline 
\multicolumn{2}{|c|}{Full routing by direct optimisation}\tabularnewline
\hline 
Decision point & Result\tabularnewline
\hline 
\hline
\textcolor{gray}{1) Overall Value} & \textcolor{gray}{High}\tabularnewline
\hline 
2) Classical hardness & Hard\tabularnewline
\hline 
3) per-instance value & Low\tabularnewline
\hline 
3a) Central server & Yes\tabularnewline
\hline 
4) Data throughput & High\tabularnewline
\hline 
\hline
\multirow{2}{*}{Conclusion} & Q-inspired\tabularnewline
 & or Reformulate\tabularnewline
\hline 
\end{tabular}

\caption{\label{tab:full_direct_opt} Path through the flowchart in Fig.~\ref{fig:design_flowchart} which leads to the decision that the full routing method of implementation 1 is probably not feasible for near term quantum computing.}
\end{table}

\subsection*{Deployment 2: Routing sub-problems by direct optimisation}

Reformulating a bit from the previous implementation, let us instead consider an implementation which rather than constructing full routing plans by direct optimisation, smaller sub-problems are passed to the quantum computer, for instance regions where routing can be particularly problematic, or where operators are finding it difficult to decide a good routing plan. 

For this deployment, points 1-3 of the flowchart are essentially the same as the previous use case. However, the data throughput to the quantum machine will be much less. Furthermore, as we have discussed previously, these optimisation problems are of a form which can be mapped directly to a quantum computer, therefore this implementation is worth exploring on a fully quantum computer in the near term. The path through the flowchart is summarized in table \ref{tab:subprob_opt}.

\begin{table}
\begin{tabular}{|c|c|}
\hline 
\multicolumn{2}{|c|}{Sub-problems by direct optimisation}\tabularnewline
\hline 
Decision point & Result\tabularnewline
\hline 
\hline
\textcolor{gray}{1) Overall Value} & \textcolor{gray}{High}\tabularnewline
\hline 
2) Classical hardness & Hard\tabularnewline
\hline 
3) per-instance value & Low\tabularnewline
\hline 
3a) Central server & Yes\tabularnewline
\hline 
4) Data throughput & Low\tabularnewline
\hline 
5) Current methods & Yes\tabularnewline
\hline 
\hline
Conclusion & Quantum\tabularnewline
\hline 
\end{tabular}

\caption{\label{tab:subprob_opt} Path through the flowchart in Fig.~\ref{fig:design_flowchart} which leads to the decision that the full routing method of implementation 2 is probably feasible for near term quantum computing.}
\end{table}

\subsection*{Deployment 3: Conditional sampling to improve dispatch policies}

Now that we have found one sensible way to implement quantum computing to solve the ambulance dispatch problem, let us examine a completely different implementation. Rather than solving the problem of dispatching ambulances directly, let us consider developing policies for ambulance dispatchers, for instance sending the second closest asset to respond to an incident if sending the first closest would leave a `hole' in the asset coverage. The Ising models which are typically used to map to quantum computers can be treated as probabilistic graphical models \cite{KollerBook} and used to perform statistical inference (a proof-of-principle application of quantum computing to error correction through probabilistic inference can be found here \cite{chancellor16b}). For the purposes of this paper, we will not discuss the details of how these models can be constructed, but it is important to note that such a construction is possible.

However, determining which, if any, of such policies to implement requires data about failures of the ambulance dispatch system. Since the system (hopefully) fails rather rarely, such data will not be readily available. This is where conditional sampling as discussed in the second approach comes in. The probabilistic graphical models which quantum computers can be used to implement, can sample based on what conditions most likely lead to a failure and this can inform policy design.

\begin{table}
\begin{tabular}{|c|c|}
\hline 
\multicolumn{2}{|c|}{Non-conditional sampling}\tabularnewline
\hline 
Decision point & Result\tabularnewline
\hline 
\hline
\textcolor{gray}{1) Overall Value} & \textcolor{gray}{High}\tabularnewline
\hline 
2) Classical hardness & Hard? \tabularnewline
\hline 
2a) Better Classical & Yes \tabularnewline
\hline
2) Classical hardness & Easy \tabularnewline
\hline
2b) Other hard subprobs & Yes \tabularnewline
\hline 
\hline
Conclusion & Reformulate\tabularnewline
\hline 
\end{tabular}

\caption{\label{tab:noncond_samp} Path through the flowchart in Fig.~\ref{fig:design_flowchart} which leads to the decision that neither quantum nor quantum inspired algorithms make sense for sampling when not conditioned on a rare event.}
\end{table}

Looking first at point 2, we first note that in general conditional sampling a problem formulated as an Ising model is certainly going to be hard to solve classically. The next question which arises is whether some other method could sample effectively, for non-conditional sampling, a simple Monte Carlo simulation based on the same probability used in the probabilistic graphical model could be constructed. However, such a Monte Carlo simulation would not preferentially sample toward meeting a given set of `failure' conditions, in other words it would not be conditional sampling. For relatively common events, just running a Monte Carlo simulation and discarding the runs where the condition is not met would suffice, however, since getting ambulance dispatch right is a fairly high stakes scenario, even modelling rare events is important, so the use of more sophisticated quantum techniques is justified from this perspective. The path through the flowchart for sampling which is \emph{not} conditioned on rare events is summarized in table \ref{tab:noncond_samp}. 

Since the sampling will be done offline, rather than directly as part of the dispatch protocol, point 3 is also satisfied. As for point 4, even conditional sampling on small sub-problems could provide useful information to inform policies, so implementing with a relatively low data throughput does make sense. Finally, as previously discussed, these problems are mappable to Ising models, so exploring quantum computing techniques is again appropriate. The path through the flowchart for sampling when conditioned on rare events is summarized in table \ref{tab:cond_samp}.

\begin{table}
\begin{tabular}{|c|c|}
\hline 
\multicolumn{2}{|c|}{Conditional sampling}\tabularnewline
\hline 
Decision point & Result\tabularnewline
\hline 
\hline
\textcolor{gray}{1) Overall Value} & \textcolor{gray}{High}\tabularnewline
\hline 
2) Classical hardness & Hard \tabularnewline
\hline 
3) per-instance value & High\tabularnewline
\hline 
4) Data throughput & Low\tabularnewline
\hline 
5) Current methods & Yes\tabularnewline
\hline
Conclusion & Quantum\tabularnewline
\hline 
\end{tabular}

\caption{\label{tab:cond_samp} Path through the flowchart in Fig.~\ref{fig:design_flowchart} which leads to the decision that conditional sampling conditioned on rare evens as described in implementation 3 is appropriate for near term quantum computation.}
\end{table}

\section{Discussion}

In this document we have laid out an objective decision making framework for use cases for quantum and quantum inspired algorithms. This includes both a simplified framework for deciding if a particular deployment of quantum computing should be investigated, and a more complex iterative methodology for designing new deployment strategies.

Both of these strategies are important and solve different problems. The first set of techniques solves a problem we have called the ``deployment decision problem'' which is the problem of deciding if a given deployment of quantum computing techniques and resources is sensible, and suggests some further courses of action. Solving this problem is useful when evaluating a fixed proposal for deploying quantum computers; this could be useful for example when evaluating a proposal for a deployment strategy critically. The problem which the second strategy framework addresses is the ``deployment design and decision problem'' which is the problem of iteratively developing a deployment of quantum  or quantum inspired computing; this more involved iterative protocol is meant for those who wish to investigate quantum and quantum inspired computing for their own use case, and are willing to iteratively improve on these methods.

With quantum computing emerging as a useful tool for industry and other end users, it is crucial to develop objective decision making methodologies which can be used without a detailed understanding of the underlying quantum computing techniques. Standardized methodologies, such as those proposed here serve the important function of increasing objectivity. Without a standard set of questions to ask, and a standard protocol to follow based on the answers to those questions, it is easy to focus on the positives of a given deployment without critically examining potential flaws and bottlenecks, or to focus too strongly on the technicalities of mapping problems before asking bigger questions about whether a use case is actually suited for quantum computing. These standard protocols also provide structure to discussions between quantum computing experts and end users, by highlighting important questions.

While we have included what we believe are the key high level questions to ask, the question of how to make decisions about quantum computing use cases and deployments is not one which can be answered in one paper, and this is not our intention - rather our intention is to start the conversation about how these decisions should be made. As the field evolves over time, it is likely that the important questions which need to be asked, and the decisions which should be made based on those answers will change, and the methodology will need to evolve with it. 

\section{Acknowledgements}

NC was supported by UK Engineering and Physical Sciences Research Council fellowship number EP/S00114X/1 and impact acceleration money associated with EP/L022303/1. TT would like to thank South Central Ambulance Service NHS Foundation Trust, Bicester, UK for their assistance and Chris Bradley of 2020 Delivery for the suggestion of ambulance dispatch as an interesting problem to consider.

\bibliography{bibLibrary}  

\end{document}